\documentclass{article}

\usepackage[preprint]{neurips_2024}

\usepackage[utf8]{inputenc} % allow utf-8 input
\usepackage[T1]{fontenc}    % use 8-bit T1 fonts
\usepackage{hyperref}       % hyperlinks
\usepackage{url}            % simple URL typesetting
\usepackage{booktabs}       % professional-quality tables
\usepackage{amsfonts}       % blackboard math symbols
\usepackage{nicefrac}       % compact symbols for 1/2, etc.
\usepackage{microtype}      % microtypography
\usepackage{xcolor}         % colors
\usepackage[namelimits]{amsmath}
\usepackage{amssymb}
\usepackage{graphicx}
\usepackage{subcaption}
\usepackage{enumitem}
\usepackage{wrapfig}
\usepackage{soul}
\usepackage{tabularx}
\usepackage{colortbl}
\usepackage[numbers]{natbib}

\title{GaussianSR: 3D Gaussian Super-Resolution with 2D Diffusion Priors}

\author{%
  Xiqian Yu 
 \textsuperscript{$^{\ast}$ 1}
    , Hanxin Zhu \textsuperscript{$^{\ast}$ 1}, Tianyu He \textsuperscript{2}, Zhibo Chen \textsuperscript{1}\\
  \textsuperscript{1} University of Science and Technology of China, \textsuperscript{2} Microsoft Research Asia\\
\texttt{\{yuxiqian,hanxinzhu\}@mail.ustc.edu.cn} \\
\texttt{tianyuhe@microsoft.com} \\
\texttt{chenzhibo@ustc.edu.cn} \\
}

\begin{document}

\maketitle

\renewcommand{\thefootnote}{\fnsymbol{footnote}}
\footnotetext[1]{Equal contribution.}
\renewcommand{\thefootnote}{\arabic{footnote}}

\begin{abstract}
 Achieving high-resolution novel view synthesis (HRNVS) from low-resolution input views is a challenging task due to the lack of high-resolution data. Previous methods optimize high-resolution Neural Radiance Field (NeRF) from low-resolution input views but suffer from slow rendering speed. In this work, we base our method on 3D Gaussian Splatting (3DGS) due to its capability of producing high-quality images at a faster rendering speed. To alleviate the shortage of data for higher-resolution synthesis, we propose to leverage off-the-shelf 2D diffusion priors by distilling the 2D knowledge into 3D with Score Distillation Sampling (SDS). Nevertheless, applying SDS directly to Gaussian-based 3D super-resolution leads to undesirable and redundant 3D Gaussian primitives, due to the randomness brought by generative priors.
 To mitigate this issue, we introduce two simple yet effective techniques to reduce stochastic disturbances introduced by SDS. Specifically, we 1) shrink the range of diffusion timestep in SDS with an annealing strategy; 2) randomly discard redundant Gaussian primitives during densification.
 Extensive experiments have demonstrated that our proposed GaussainSR can attain high-quality results for HRNVS with only low-resolution inputs on both synthetic and real-world datasets. Project page: \url{https://chchnii.github.io/GaussianSR/}.

\end{abstract}

\section{Introduction}

Novel View Synthesis (NVS) has been extensively studied in computer vision and graphics. In particular, Neural Radiance Field (NeRF)~\cite{mildenhall2021nerf} has demonstrated its impressive ability to generate high-quality visual content. More recently, 3D Gaussian Splatting (3DGS)~\cite{kerbl3Dgaussians} has been attracting widespread attention due to its capability of producing high-quality images with faster rendering speed. However, achieving high-resolution novel view synthesis (HRNVS) from low-resolution inputs remains an under-explored yet challenging task. 

There exist two primary difficulties for HRNVS. Firstly, previous works~\cite{wang2022nerf, han2023super, huang2023refsr} mainly rely on optimizing high-resolution NeRF from low-resolution input views. Although these methods can synthesize satisfactory high-resolution novel views, the stratified sampling required for rendering in NeRF is costly and results in high rendering time. Secondly, we only have low-resolution input views to produce high-resolution results. To tackle this, NeRF-SR~\cite{wang2022nerf} exploits a supersampling strategy to estimate color and density at the sub-pixel level. However, it is still a challenge to get enough information from the low-resolution input alone.

In this work, we propose GaussianSR, which aims to introduce 2D generative priors learned from large-scale image data into HRNVS. Specifically, we build our method upon 3DGS due to its photorealistic visual quality and real-time rendering. In order to leverage 2D priors, we derive inspiration from DreamFusion~\cite{poole2022dreamfusion}, a method that distills 2D diffusion priors into text-to-3D generation with Score Distillation Sampling (SDS). In this way, to introduce 2D priors to HRNVS, a straightforward solution is to distill off-the-shelf 2D super-resolution diffusion priors into 3DGS for high-resolution novel view synthesis. Nevertheless, we notice that applying SDS directly fails with some undesirable and redundant 3D Gaussian primitives. We suspect that this is due to the inherent randomness of the generative priors, as it always takes random noise and timestep as input to produce natural image distribution. This property is particularly amplified in SDS when we aim to optimize high-resolution 3DGS with denser Gaussian primitives, since there are large variances in the gradients during 3DGS densification (a process that clones or splits current Gaussian primitives into more). To mitigate this issue, we propose two simple yet effective techniques, which reduce stochastic disturbances introduced by SDS. Firstly, to alleviate the randomness of the diffusion timestep, we shrink the sampling range of the diffusion timestep with an annealing strategy. Secondly, to prevent explosive Gaussian primitives, we randomly discard redundant primitives during the process of densification.
We validate our GaussianSR in various scenarios, including synthesized and realistic scenarios, and experimental results demonstrate that the rendering quality of GaussianSR outperforms existing state-of-the-art methods.

In conclusion, our contributions can be summarized as follows:
\begin{itemize}[leftmargin=*]
\item To alleviate the lack of high-resolution data, we, for the first time, propose to distill generative priors of 2D super-resolution models into HRNVS.

\item We observe that applying SDS directly to Gaussian-based 3D super-resolution leads to undesirable and redundant 3D Gaussian primitives, due to randomness brought by generative priors. To solve this issue, we propose two techniques to reduce stochastic disturbances introduced by SDS.

\item  Experimental results demonstrate that our proposed GaussianSR achieves higher-quality HRNVS than the state-of-the-art solutions from only low-resolution inputs.

\end{itemize}

\section{Related Work}

\subsection{3D Gaussian Splatting}

3D Gaussian Splatting (3DGS) \cite{kerbl3Dgaussians} provides a promising and effective approach for Novel View Synthesis (NVS). By representing the scene explicitly as a collection of 3D Gaussian primitives and rendering views through rasterization, 3DGS achieves impressive quality and speed in NVS. 
This has led to the emergence of various 3DGS extensions, including modeling dynamic scenes \cite{wu20234d, yang2023real,lu20243d,huang2023sc}, reconstruction without input camera poses \cite{fu2023colmap,fan2024instantsplat}, few-shot view synthesis \cite{zhu2023fsgs,chung2023depth,li2024dngaussian} and applications in other fields \cite{zhou2023drivinggaussian,tang2023dreamgaussian,liang2023luciddreamer,chung2023luciddreamer,tang2023make}.
Additionally, some studies \cite{huang2024gs++,bulo2024revising,yang2024spec,cheng2024gaussianpro} focus on advancing 3DGS itself to improve the quality of NVS.
However, existing relevant works 
have primarily concentrated on synthesizing novel views with resolutions limited to the input views, neglecting the high-resolution novel views synthesis (HRNVS) from low-resolution inputs.
In this paper, we explore high-quality HRNVS with 3DGS.

\begin{figure}[t]
    \centering
    \includegraphics[width=1\textwidth]{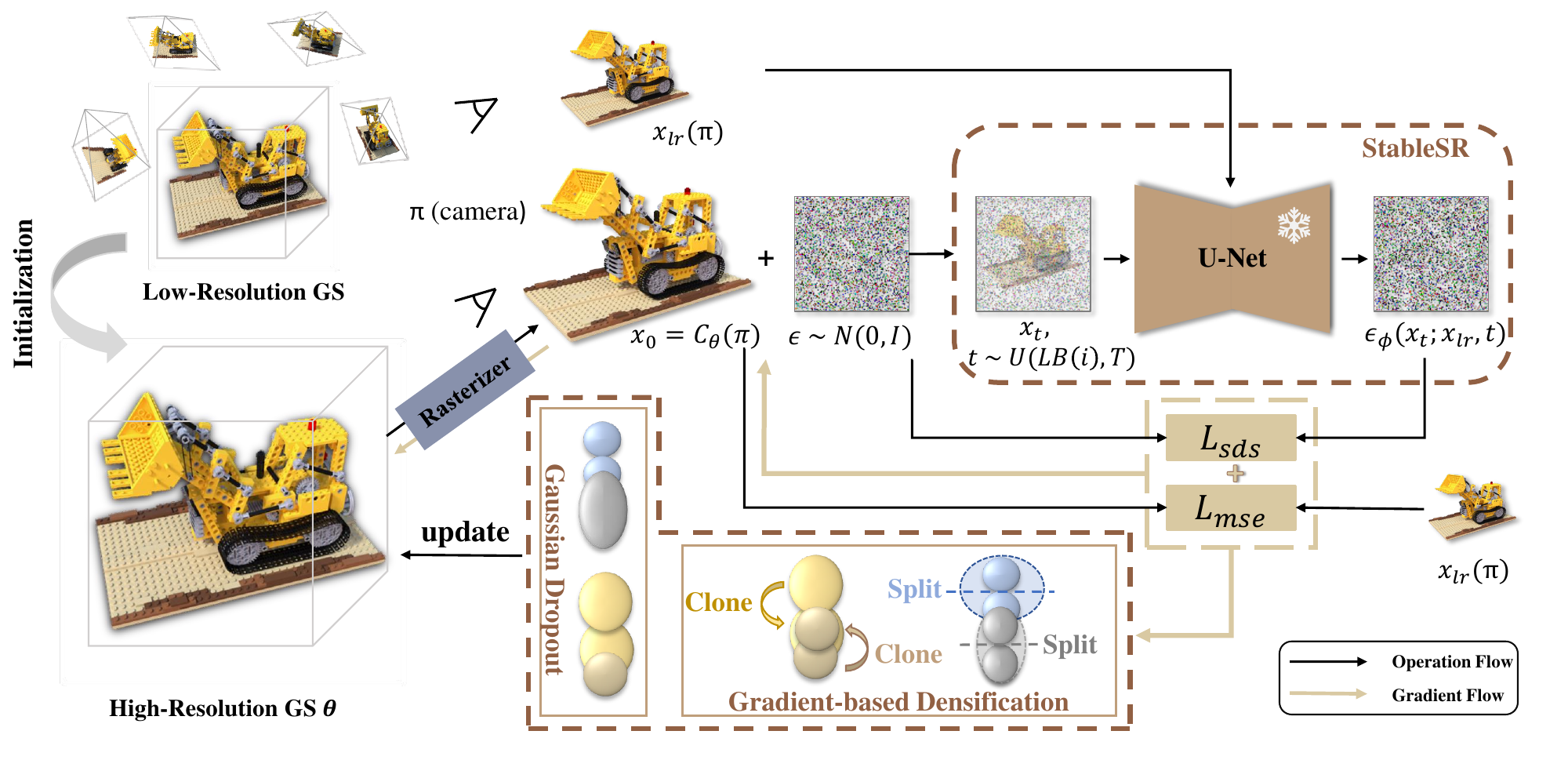}
    \caption{Overview of GaussianSR. To alleviate the lack of high-resolution data, we synthesize high-resolution novel views by distilling 2D diffusion priors into 3D representation with SDS (Sec.~\ref{3.1}). Since the redundant Gaussian primitives are introduced due to the randomness of generative priors (Sec.~\ref{3.2}), we propose Gaussian Dropout and diffusion timestep annealing to reduce stochastic disturbance (Sec.~\ref{3.3}).}
    \label{framework}
\end{figure}
\subsection{High-Resolution Novel View Synthesis}
High-resolution novel view synthesis (HRNVS) aims to synthesize high-resolution novel views from only low-resolution inputs. As a pioneer, NeRF-SR \cite{wang2022nerf} optimizes high-resolution NeRF with the sub-pixel constraint, ensuring that the values of low-resolution (LR) pixels equal the mean value of high-resolution (HR) sub-pixels
RefSR-NeRF \cite{huang2023refsr} reconstructs the high-frequency details with the help of a single high-resolution reference image. Furthermore, Super-NeRF \cite{han2023super} constructs a consistency-controlling super-resolution module to generate view-consistent high-resolution details for NeRF. However, these NeRF-based methods suffer from slow rendering speed. Recently, 3DGS has gained popularity due to its primitive-based representation, which can produce high-quality images at faster rendering speeds. While a concurrent work, SRGS \cite{feng2024srgs}, similarly focuses on 3DGS-based HRNVS, however, our study and SRGS differ significantly not only in terms of technical contributions but also in motivation: we aim to introduce 2D diffusion priors, which is learned from large-scale 2D data, into HRNVS.

\section{Methodology}
In this section, we provide a comprehensive overview of our proposed GaussianSR. To begin with, recognizing the challenge posed by the limit availability of high-resolution data, we leverage 2D diffusion priors distilled by SDS~\cite{poole2022dreamfusion} to optimize high-resolution 3DGS (Sec.~\ref{3.1}). However, the randomness introduced by generative priors will lead to undesirable and redundant Gaussian primitives (Sec.~\ref{3.2}). Hence, we propose two simple yet effective techniques to mitigate this issue (Sec.~\ref{3.3}).

\subsection{3DGS Super-Resolution with SDS Optimization}
\label{3.1}

\paragraph{3DGS.}
\label{3.1.1}
As an effective method for novel view synthesis, 3D Gaussian Splatting (3DGS)~\cite{kerbl3Dgaussians} represents a 3D scene using a series of Gaussian primitives comprised of position $\boldsymbol \mu \in \mathbb{R}^3$, scaling $\boldsymbol s \in \mathbb{R}^3$, rotation $\boldsymbol r \in \mathbb{R}^3$, and color $\boldsymbol c \in \mathbb{R}^3$. To faithfully reconstruct the 3D scene, these Gaussian primitives are initialized with sparse point clouds estimated by SfM~\cite{schonberger2016structure}, followed by a densification operation, \textit{i.e}., the split and clone operation,  that adaptively control their numbers and densities.
Concretely,
whether a 3D Gaussian primitive is split or cloned is determined by the average gradient magnitude of the Normalized Device Coordinates (NDC) \cite{mcreynolds2005advanced} for the viewpoints in which the Gaussian primitive participates in the calculation. For example, for Gaussian primitive $k$ under viewpoint $M_i$, the NDC is $(\mu^{k,M_i}_{ndc,x},\mu^{k,M_i}_{ndc,y},\mu^{k,M_i}_{ndc,z})$, and the loss under viewpoint $M_i$ is $\mathcal{L}_{M_i}$. During optimization, Gaussian primitive $k$ participates in the calculation for $M$ viewpoints. When Gaussian primitive satisfies:
\begin{equation}
    \begin{aligned}
        |g|=
        \frac{\sum_{M_i =1}^{M}\sqrt{\left(\frac{\partial\mathcal{L}_{M_i}}{\partial \mu^{k,M_i}_{ndc,x}}\right)^2 + \left(\frac{\partial\mathcal{L}_{M_i}}{\partial \mu^{k,M_i}_{ndc,y}}\right)^2}}{M} > \tau_{pos},
    \end{aligned}
    \label{eq1}
\end{equation}
it is transformed into two Gaussian primitives, where $\tau_{pos}$ is the default threshold.

\paragraph{Distilling 2D Diffusion Priors for 3DGS Super-Resolution.}

The primary challenge encountered in high-resolution novel view synthesis (HRNVS) from low-resolution inputs is the scarcity of data, a limitation pervasive across various domains. For instance, in text-to-3D generation, the performance of early endeavours~\cite{sanghi2022clip,sanghi2023clip,yang2019pointflow} is limited by the small-scale text-3D datasets adopted (\textit{e.g.}, ShapeNet~\cite{chang2015shapenet}), resulting in poor generalization. To overcome the bottleneck of data scarcity and facilitate the generation of more diverse 3D assets, DreamFusion~\cite{poole2022dreamfusion} introduces Score Distillation Sampling (SDS), which aims to distill generative priors from pretrained text-to-image diffusion models. Drawing inspiration from DreamFusion, we propose leveraging  off-the-shelf 2D super-resolution diffusion priors to mitigate the data shortage challenge in HRNVS.

Specifically, as shown in Fig.~\ref{framework}, given a set of multi-view low-resolution images $x_{lr}$, our objective is to synthesize high-resolution novel views through optimizing high-resolution 3DGS with SDS.
Initially, we reconstruct a low-resolution 3DGS from the multi-view low-resolution inputs, which serves as the initialization for the high-resolution 3DGS. Subsequently, we optimize the high-resolution 3DGS using priors distilled from a diffusion-based 2D super-resolution model along with the low-resolution inputs.
Let $\boldsymbol C_{\boldsymbol \theta} (\pi)$ represent the rendered high-resolution image at the given viewpoint $\pi$, where $\boldsymbol C$ is the differentiable rendering function for the high-resolution 3DGS parameterized by $\boldsymbol \theta$. 
Our goal is to optimize the rendered high-resolution image, denoted as $x_0 := \boldsymbol C_\theta (\pi)$, by introducing the SDS loss $ \mathcal{L}_{SDS}$, 
which encourages $x_0$ move toward higher density region conditioned on its corresponding low-resolution image $x_{lr}$. 
Particularly, $\mathcal{L}_{SDS}$ computes the difference of predicted noise $\epsilon_{\phi}$ and the added noise $\epsilon$ as per-pixel gradient, which is then used to update the high-resolution 3DGS parameters $\theta$:
\begin{equation}
    \begin{aligned}
        \nabla_{\boldsymbol \theta}\mathcal{L}_{SDS}(\phi,\boldsymbol C_{\boldsymbol\theta}) = \mathbb E _{t,\epsilon} \left[w(t) (\epsilon_{\phi}(x_t; x_{lr} ,t)- \epsilon) \frac{\partial x}{\partial \boldsymbol \theta}\right],
    \end{aligned}
\end{equation}
where $\phi$ is the pretrained image super-resolution diffusion model, $x_t$ is $x_0$ add noise $\epsilon$ at different diffusion timestep $t$, and $w(t)$ is a weight function of different noise levels.

Furthermore, to maintain the structural consistency and to prevent color shifts occasionally caused by diffusion model \cite{choi2022perception}, 
the sub-pixel constraint $\mathcal{L}_{MSE}$ is also taken into consideration as a regularizer.
The rendered high-resolution image $x_0$ is downsampled to align with its corresponding low-resolution image $x_{lr}$, which is formulated as follows:
\begin{equation}
    \begin{aligned}
        \mathcal{L}_{MSE} = ||Downsample(x_0)-x_{lr}||.
    \end{aligned}
    \label{eq_l1}
\end{equation}
In conclusion, the high-resolution 3DGS is joint optimized by $\mathcal{L}_{MSE}+\lambda \mathcal{L}_{SDS}$.

\begin{figure}[t]
	
	\begin{minipage}{0.32\linewidth}
		\vspace{3pt}
        %这个图片路径替换成你的图片路径即可使用
		\centerline{\includegraphics[width=\textwidth]{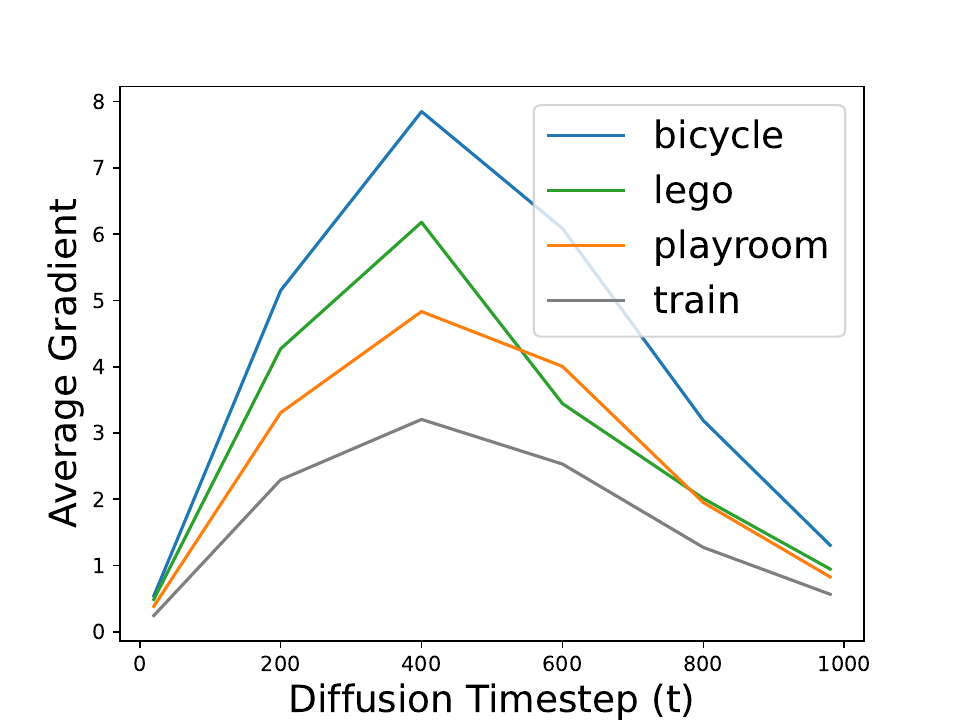}}
          % 加入对这列的图片说明
		\centerline{(a) Gradients between different $t$}
	\end{minipage}
	\begin{minipage}{0.64\linewidth}
		\vspace{3pt}
		\includegraphics[width=0.49\textwidth]{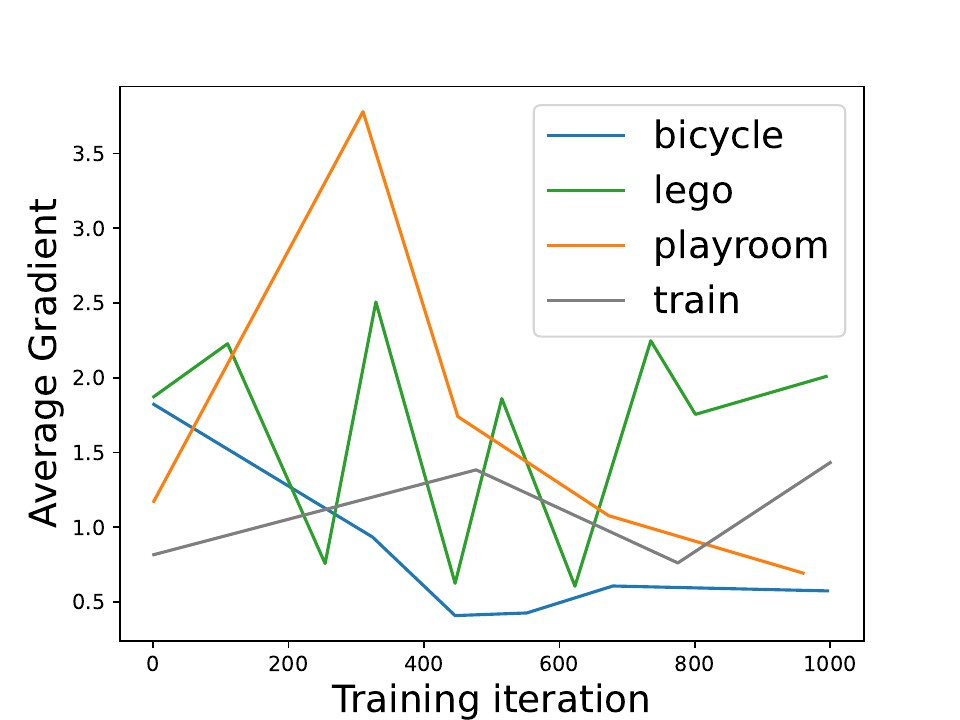}
  % \hspace{0.1in}
      \includegraphics[width=0.49\textwidth]{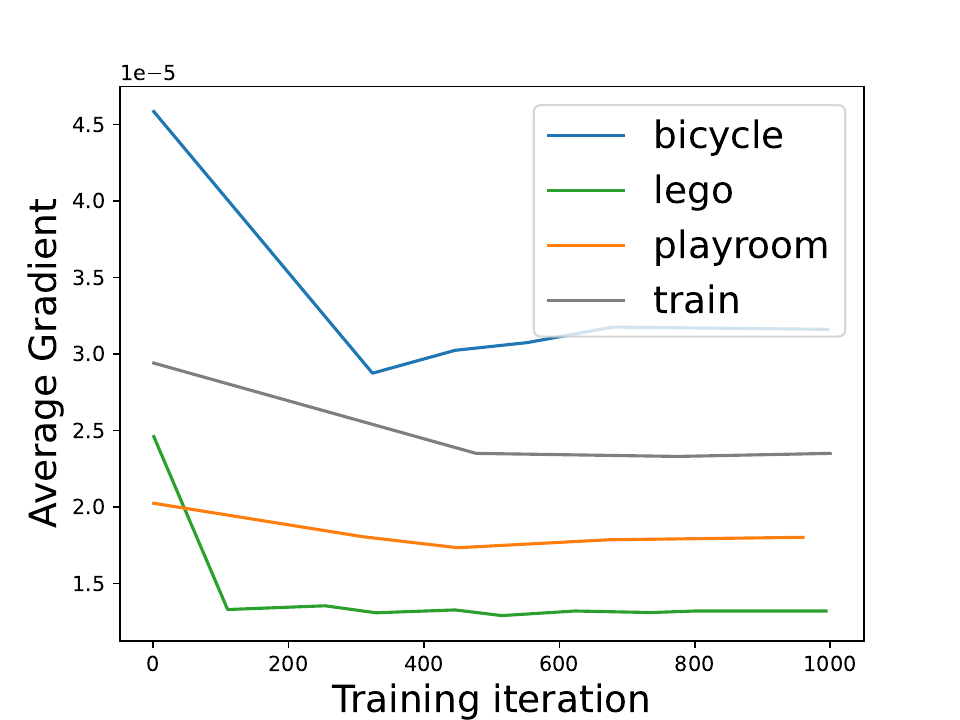}
	 
		\centerline{(b) Gradients with SDS loss (left) and MSE loss (right) }
	\end{minipage}
 
	\caption{\textbf{(a)} The gradient values under the constraint of SDS loss are visualized, revealing substantial variance across different diffusion timesteps $t$.  \textbf{(b)} When comparing the gradient values under two different constraints—SDS loss on the left and MSE loss on the right—the gradient variance for SDS is significantly larger than that for MSE.}
	\label{grad}
\end{figure}
\subsection{Gaussian Densification with SDS Constraint}
\label{3.2}
 
In 3DGS, synthesizing high-quality novel views depends significantly on the representation capacity of Gaussian primitives. In particular, achieving accurate high-resolution rendering requires denser Gaussian primitives \cite{yan2023multi}. 
In our study, denser Gaussian primitives are produced by optimizing the high-resolution 3DGS with SDS.
However, we observe that the direct application of SDS introduces undesirable and redundant Gaussian primitives during the densification process. We hypothesize that this issue arises from the inherent randomness of generative priors, as random noise and diffusion timesteps are sampled in the SDS process.

Referring to the training strategy of the diffusion model, SDS aims to optimize rendered high-resolution images to closely match the high-resolution distribution conditioned on its low-resolution counterparts by leveraging the data noising process. Nonetheless, during data noising, diffusion timesteps are randomly sampled, resulting in varying gradient values with significant variance across different iterations. As described in Sec.~\ref{3.1.1}, the Gaussian primitives with gradients exceeding a default threshold are transformed into two Gaussian primitives during densification. Consequently, the substantial variance in gradient values introduced by SDS leads to the generation of redundant Gaussian primitives.

Specifically, Fig.~\ref{grad} (a) visualizes the gradient values across different diffusion timesteps $t$. As diffusion timesteps are randomly sampled in each training iteration, the significant variance of gradient values persists throughout the training process.
Furthermore, we also visualize the variation of gradient values for a specific view during training under different constraints. The left figure in Fig.~\ref{grad} (b) shows the gradient values under $\mathcal{L}_{SDS}$, which presents a large variance compared to the right figure under $\mathcal{L}_{MSE}$. Notably, the original 3DGS \cite{kerbl3Dgaussians} employs $\mathcal{L}_{MSE}$ as the optimization constraint, which exhibits small variance across iterations and is well-suited to the default threshold strategy.
In contrast, in our study, the high-variance gradient values brought by SDS,
when subjected to the default threshold, can lead to the generation of redundant Gaussian primitives.

\begin{figure}[t]
    \centering
    \includegraphics[width=1\textwidth]{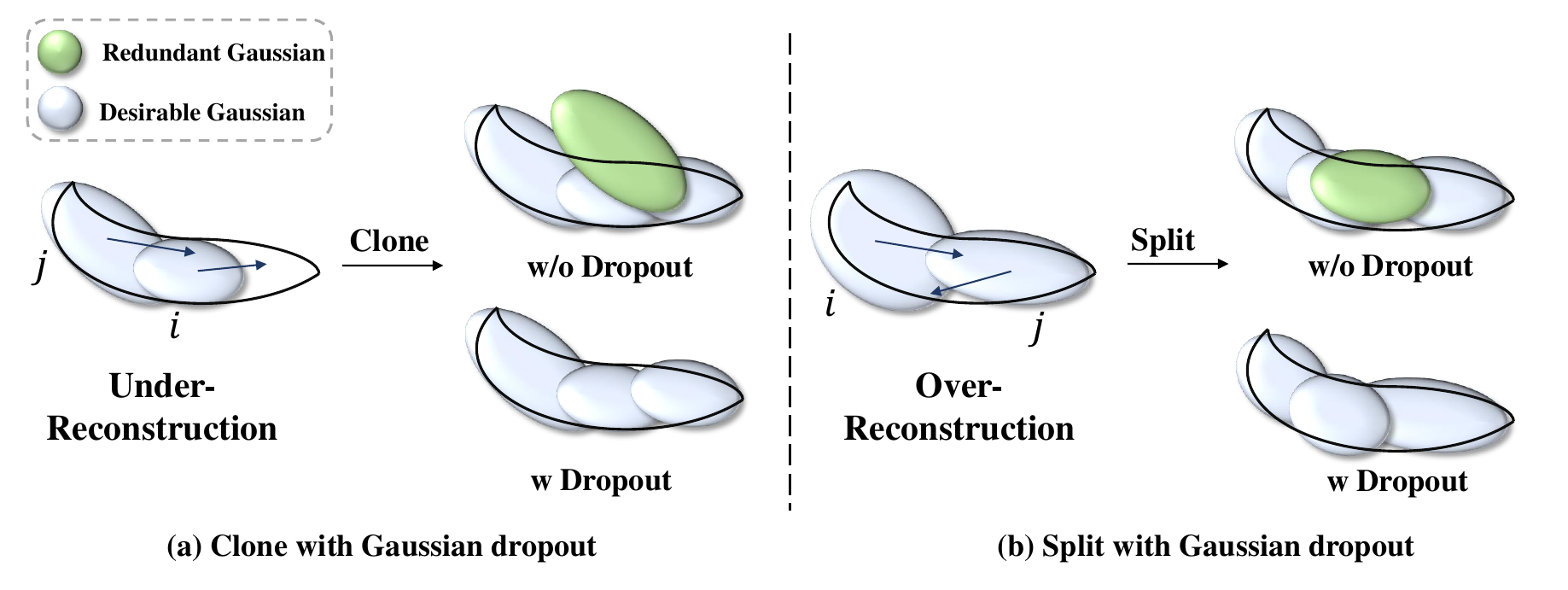}
    \caption{Illustration of Gaussian Dropout during the densification process. When a small-scale object (depicted by the black outline) is insufficiently covered (under-reconstructed) or is represented by overly large splats (over-reconstructed), cloning or splitting is performed. In the top row (without dropout), a redundant Gaussian primitive (shown in green) is generated during densification. In the bottom row (with dropout), the redundant Gaussian primitive is randomly discarded.}
    \label{dropout}
\end{figure}

\subsection{Stochastic Disturbance Reduction}
\label{3.3}
To mitigate the aforementioned problem, we propose two techniques to reduce stochastic disturbances introduced by SDS:  shrinking the range of diffusion timestep with an annealing strategy, and randomly discarding redundant Gaussian primitives during densification.

\paragraph{Diffusion Timestep Annealing.}
As a class of score-based generative models~\cite{ho2020denoising,song2020denoising,song2019generative,song2020score}, diffusion models involve a data noising and denoising process according to a predefined schedule over a fixed number of timesteps. Analogous to the training strategy of DDPM \cite{ho2020denoising}, the vanilla SDS randomly samples diffusion timestep $t$ from a uniform distribution (\textit{i.e.}, $t \sim \mathcal{U}(1,T)$) throughout the 3D model optimization. As described in Sec.~\ref{3.2}, random sampling of 
diffusion timestep $t$ in SDS leads to redundant Gaussian primitives during the densification process. Therefore, we revise the timestep sampling range in SDS with an annealing strategy to reduce stochastic disturbances.

Particularly, the vanilla timestep sampling strategy of SDS involves sampling $t$ from a fixed range $[1, T]$ during each data noising step. In our approach, we refine it by employing an annealing strategy to progressively shrink the lower bound of the diffusion timestep sampling range.
Specifically, for the current iteration $i$, the sampling range is adjusted to $[LB(i), T]$, where the lower bound $LB(i)$ is calculated as follows:
% In detail, the sampling range is set to $[LB(i),T]$ for the current iteration $i$, where the lower bound of sampling range can be calculated by:
\begin{equation}
    \begin{aligned}
        LB(i) = T - \frac{i}{N}.
    \end{aligned}
    \label{eq_t}
\end{equation}
In this equation, $T$ represents the upper bound, and $N$ denotes the annealing interval. Consequently, the diffusion timestep $t$ is sampled from the interval $[LB(i), T]$, i.e., $t \sim \mathcal{U}(LB(i), T)$, during each data noising step in the SDS process.

\paragraph{Gaussian Dropout.}
In addition to reducing stochastic disturbances by shrinking the diffusion timestep sampling range, we directly discard undesirable and redundant Gaussian primitives using Gaussian Dropout.
Specifically, as depicted in Fig.~\ref{dropout} (a), when considering the cloning of Gaussian primitive $i$ to fill the empty area (referred to as the "under-reconstruction" region), the nearby Gaussian primitive $j$, which should not be cloned, may exhibit large gradients due to disturbances from the SDS loss. This can lead to the generation of redundant Gaussian primitives. Therefore, to mitigate this issue, we employ Gaussian Dropout to discard the cloning or splitting of certain Gaussian primitives.

In detail, given a set of Gaussian primitives $G = \{g_0, g_1,..., g_n\} \in \mathbb{R}^n $, where all gradients exceed the default threshold $\tau_{pos}$, we first generate a mask $M$ randomly with a certain probability $p$ and then use the mask to select a subset $G^{'} = \{g_0, g_2,...,g_{n-2}, g_n\} \in \mathbb{R}^k (k<n)$ of $G$. The Gaussian primitives in subset $G^{'}$ will be split or cloned during densification, while the other Gaussian primitives will be dropped out and remain unchanged.
Thus, the denser set $\hat{G} \in \mathbb{R}^{n+k}$ after densification can be formulated as:
\begin{equation}
    \begin{aligned}
        \hat{G} = \mathcal{D}(G^{'}) + (G - G^{'}),
        \ \
        \text{where}
        \ 
        G^{'} = G \cdot M(p),
        M(p) = 
        \begin{cases}
            0 & rand(G) < p\\
            1 & else \\
        \end{cases},
    \end{aligned}
    \label{eq_dropout}
\end{equation}
where $\mathcal{D}$ means the densification step.

\section{Experiments}
In this section, we present a comprehensive set of qualitative and quantitative evaluations aimed to verify the effectiveness of our proposed GaussianSR. Additionally, we conduct ablation studies to systematically evaluate the impact and effectiveness of each individual component.

\subsection{Datasets and Metrics}

\paragraph{Blender Dataset \cite{mildenhall2021nerf}.}
Blender Dataset is a Realistic Synthetic $360^\circ$ Dataset that contains 8 detailed synthetic objects with a resolution of $800\times800$. We follow the same training and testing data split strategy as the original 3DGS \cite{kerbl3Dgaussians}. For each scene, 100 images are used for training and 200 images are used for testing. The input resolution is set to $200 \times 200$, and we super-resolve this low-resolution 3DGS by a factor of 4. The downsampling method used is the same as the one provided in the official 3DGS code.

\paragraph{Mip-NeRF 360 Dataset \cite{barron2022mip}.}
Mip-NeRF 360 consists of 9 real-world scenes with 5 outdoors and 4 indoors. Each of them is composed of a complex central object or area with a detailed background. Following the previous setup, we use $7/8$ of the images for training and take the remaining $1/8$ for testing in each scene. We 
downsample the training views by a factor $4$ as low-resolution inputs to $\times4$ HRNVS task.

\paragraph{Deep Blending Dataset \cite{hedman2018deep}.}
Deep Blending is a real-world dataset. Following 3DGS~\cite{kerbl3Dgaussians}, we select two scenes of Deep Blending to evaluate our method. We use 1/8 of all views for testing and the rest for training.  We downsample the training views by a factor $4$ as low-resolution inputs to $\times4$ HRNVS task.

\paragraph{Metrics.}
The quality of view synthesis is assessed relative to the ground truth from the same pose, employing four metrics:
Peak Signal-to-Noise Ratio (PSNR) and Structural Similarity Index Measure (SSIM) \cite{wang2003multiscale}, LPIPS (VGG) \cite{zhang2018unreasonable} and Frames Per Second (FPS).

\begin{table}[t]
\scriptsize
\centering
\caption{Quantitative comparison for HRNVS ($\times4$) with previous works on Blender, Mip-NeRF 360, and Deep Blending Dataset.}
\begin{tabularx}{\textwidth}{l|*{4}{>{\centering\arraybackslash}X}|*{4}{>{\centering\arraybackslash}X}|*{4}{>{\centering\arraybackslash}X}}

 & \multicolumn{4}{c|}{Blender} & \multicolumn{4}{c|}{Mip-NeRF 360} & \multicolumn{4}{c}{Deep Blending} \\

 Method & PSNR$\uparrow$ & SSIM$\uparrow$ & LPIPS$\downarrow$ & FPS$\uparrow$ & PSNR$\uparrow$ & SSIM$\uparrow$ & LPIPS$\downarrow$ & FPS$\uparrow$ & PSNR$\uparrow$ & SSIM$\uparrow$ & LPIPS$\downarrow$ & FPS$\uparrow$\\
\hline
3DGS \cite{kerbl3Dgaussians} & 21.78
 & 0.868 & 0.104& \textbf{192} & 20.28 & 0.581 & 0.420& \textbf{33} & 26.64 & 0.854 & 0.312 & \textbf{60} \\

StableSR \cite{wang2023exploiting} & 23.57 & 0.854 & 0.207 &  <1& 21.83 & 0.467 & 0.383&  <1& 23.93 & 0.708 & 0.325& <1\\

Bicubic & 27.23 & 0.911 & 0.115 & 93 & 25.14 & 0.618 & 0.406& 27& 28.01 & 0.864 & 0.330 & 40\\

NeRF-SR \cite{wang2022nerf} & 27.81 & 0.920 & 0.097& <1 & -- & -- & -- & -- & -- & -- & -- & -- \\

Ours & \textbf{28.37} &\textbf{0.924} & \textbf{0.087} &\textbf{192} & \textbf{25.60} & \textbf{0.663} &\textbf{0.368}  &  \textbf{33} &\textbf{28.28}&\textbf{0.873}  &\textbf{0.307} & \textbf{60}\\

% \hline
\end{tabularx}
\label{table_main_results}
\end{table}

\begin{figure}[t]
    \centering
    \includegraphics[width=1\textwidth]{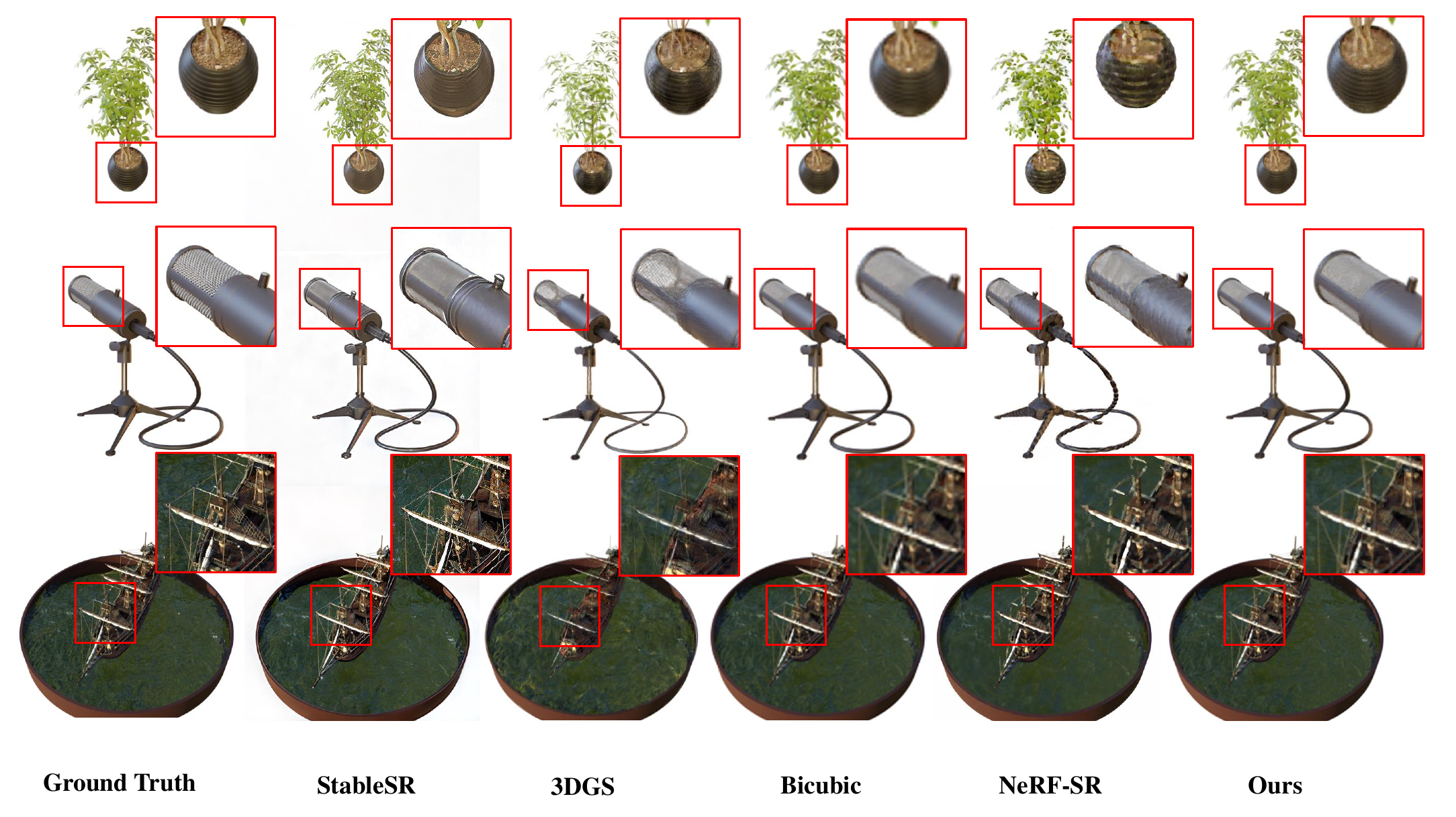}
    \caption{Qualitative comparison of the HRNVS ($\times4$) on Blender dataset. Our method shows clearer details than 3DGS~\cite{kerbl3Dgaussians}, Bicubic, NeRF-SR~\cite{wang2022nerf} and StableSR~\cite{wang2023exploiting}. }
    \label{main_results_blender}
\end{figure}

\subsection{Implementation Details}
We implement our method based on the open-source 3DGS code. Training consists of 30k iterations for indoor scenes and 10k iterations for other scenes. As for the off-the-shelf 2D super-resolution diffusion model, we opt for StableSR \cite{wang2023exploiting} as our backbone. For the annealing interval $N$ in Eq.~\ref{eq_t}, we shrink the sampling range of diffusion timestep every 100 iterations. The dropout probability $p$ of 0.7 is set during the Gaussian Dropout process.
Additionally, bilinear interpolation is employed to downsample the rendered high-resolution images
for $\mathcal{L}_{MSE}$ in Eq.~\ref{eq_l1}, and $\lambda$ is set to be 0.001 during training. We perform experiments using a NVIDIA A100 GPU.
To save space, please refer to our supplementary materials for more details.

\begin{figure}[t]
    \centering
    \includegraphics[width=1\textwidth]{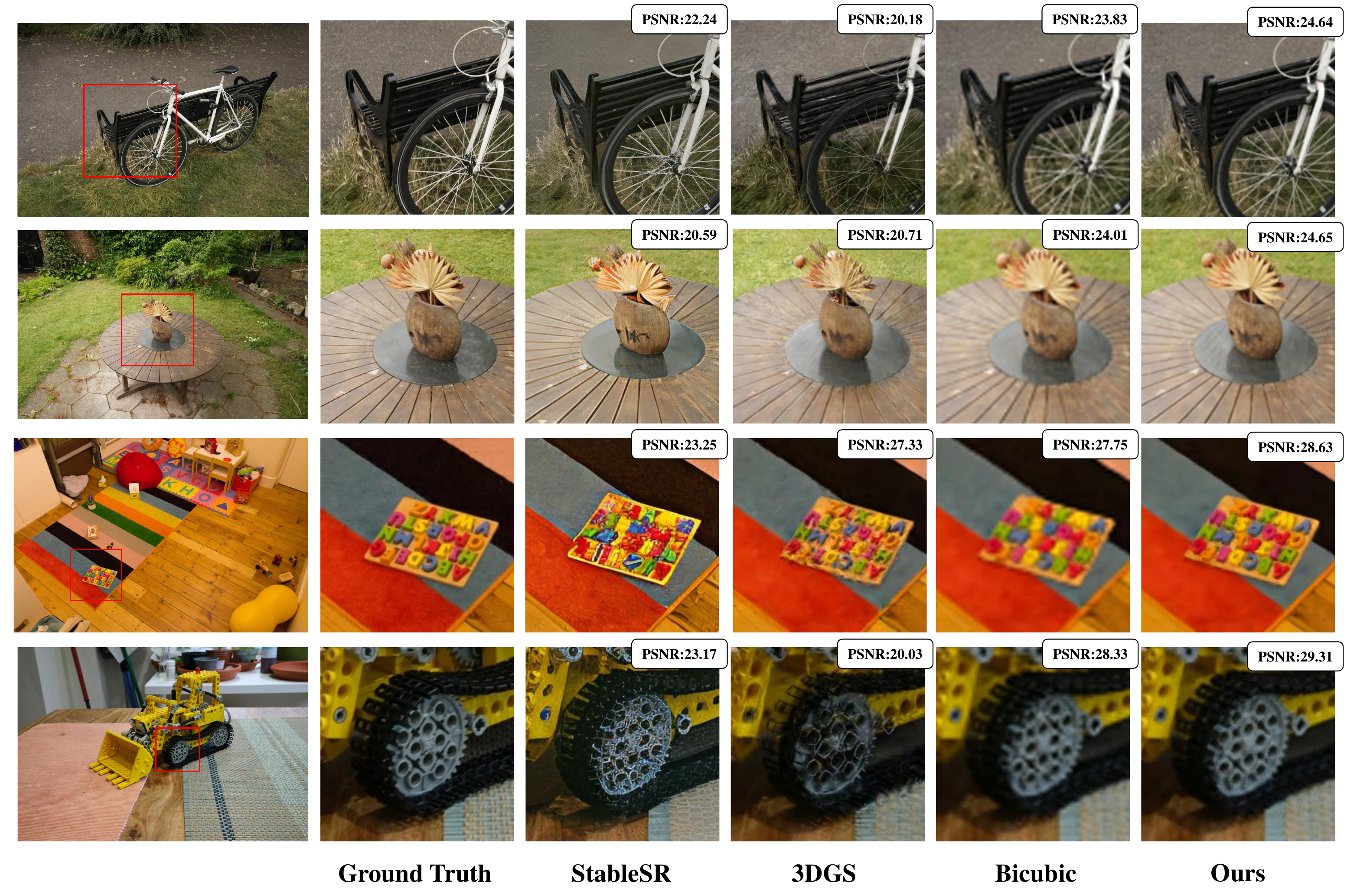}
    \caption{Qualitative comparison of our method with vanilla 3DGS, bicubic interpolation, and StableSR on Mip-NeRF 360 and Deep Blending Dataset for the HRNVS ($\times4$). The results are the zoom-in version of the red box region and the PNSR value for the current view is presented in the top right corner. Our method presents higher quality and clearer details than others.}
    \label{main_results_mip360}
\end{figure}

\subsection{Quantitative and Qualitative Comparisons}

To demonstrate the effectiveness of our method, we compare it against several prior approaches, including vanilla 3DGS \cite{kerbl3Dgaussians}, bicubic interpolation, NeRF-SR \cite{wang2022nerf} and StableSR \cite{kerbl3Dgaussians}. For vanilla 3DGS baseline, we train 3DGS \cite{kerbl3Dgaussians} using low-resolution input views and then render it at high resolution. Bicubic interpolation is applied to low-resolution renderings from the baseline 3DGS, providing a standard upsampling method.
Regarding NeRF-SR \cite{wang2022nerf}, we directly run the source code to obtain qualitative and quantitative results. 
However, due to training instabilities encountered with Mip-NeRF 360 \cite{barron2022mip} and Deep Blending Dataset \cite{hedman2018deep}, we reproduce the results of NeRF-SR only on the Blender dataset \cite{mildenhall2021nerf}.
For StableSR \cite{wang2023exploiting}, we super-resolve each low-resolution view rendered from the baseline 3DGS using StableSR. Notably, since the 2D diffusion model we adopted (\textit{i.e.}, StableSR \cite{wang2023exploiting}) is primarily trained on $4\times$ super-resolution data, we also primarily validate our GaussianSR on $\times4$ HRNVS. 

\paragraph{Quantitative Evaluation.}
Tab.~\ref{table_main_results} presents quantitative comparison results for $\times4$ HRNVS tasks on the Blender dataset \cite{mildenhall2021nerf}, the Mip-NeRF 360 dataset \cite{barron2022mip}, and the Deep Blending dataset \cite{hedman2018deep}.
Our proposed GaussianSR outperforms previous state-of-the-art methods significantly in terms of PSNR, SSIM, and LPIPS metrics, while also requiring less rendering time.
This indicates that GaussianSR excels in synthesizing high-resolution views with both superior quality and efficiency. Furthermore, our method demonstrates the capability to generate detailed high-resolution novel views solely from low-resolution inputs, across synthetic as well as real-world datasets.

\begin{table}[t]
\centering
\caption{Ablation studies on Mip-NeRF 360 and Deep Blending dataset for HRNVS (×4).}
\begin{tabularx}{\textwidth}{l|*{3}{>{\centering\arraybackslash}X}|*{3}{>{\centering\arraybackslash}X}}

 & \multicolumn{3}{c|}{Mip-NeRF 360} & \multicolumn{3}{c}{Deep Blending} \\

 Method & PSNR$\uparrow$ & SSIM$\uparrow$ & LPIPS$\downarrow$ & PSNR$\uparrow$ & SSIM$\uparrow$ & LPIPS$\downarrow$ \\
\hline
Baseline (3DGS) & 20.28 & 0.581 & 0.420 & 26.64 & 0.854 & 0.312 \\

+ $\mathcal{L}_{MSE}$ & 24.95 & 0.633 & 0.371 & 27.85 & 0.867 & 0.316  \\

+ $\mathcal{L}_{SDS}$ (w dropout) &25.36  &0.631  & 0.369 &28.18  &0.874  &0.311 \\

+ Diffusion timestep annealing &\textbf{25.60} & \textbf{0.663} &\textbf{0.368}  &\textbf{28.28}&\textbf{0.873}  &\textbf{0.307}  \\

% \hline
\end{tabularx}
\label{table_ablation}
\end{table}
\paragraph{Qualitative Evaluation.}
Fig.~\ref{main_results_blender} presents the qualitative results on the Blender dataset \cite{mildenhall2021nerf}, while
Fig.~\ref{main_results_mip360} shows the qualitative results on the Mip-NeRF 360 \cite{barron2022mip} and Deep Blending dataset \cite{hedman2018deep}. 
GaussianSR consistently exhibits high-quality visual results across various scenarios, encompassing indoor and outdoor scenes.
In contrast, the baseline model, 3DGS \cite{kerbl3Dgaussians}, suffers from needle-like artifacts due to the out-distribution rendering,
whereas bicubic interpolation yields blurring artifacts by directly interpolating low-resolution views. Results super-resolved by StableSR \cite{wang2023exploiting} directly appear coarse and suffer from color shifts. 
Across both synthesis and real-world datasets, our GaussianSR produces superior visual results characterized by clearer edges and sharper details compared to alternative methods. 

\begin{figure}
    \centering
    \includegraphics[width=1\textwidth]{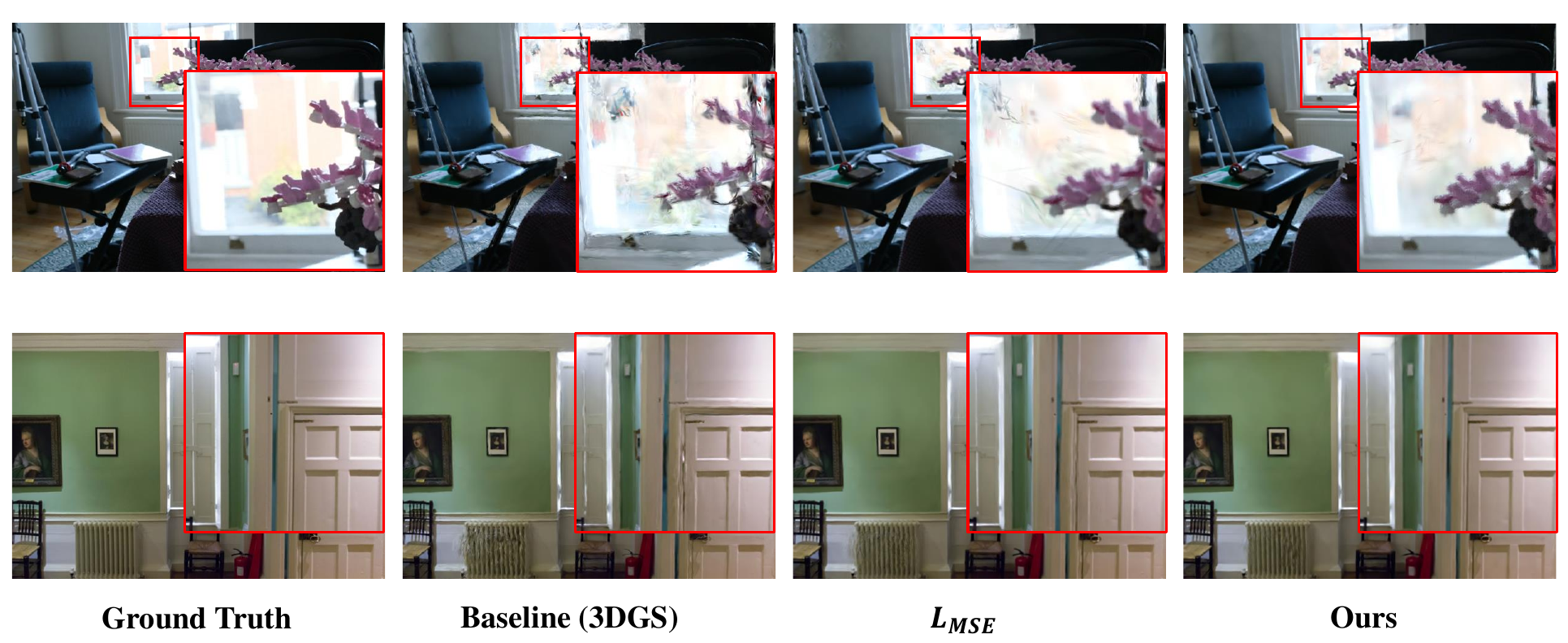}
    \caption{Qualitative evaluation for ablation studies. The third column means the results of high-resolution 3DGS that are optimized with $\mathcal{L}_{MSE}$ only. The last column presents the results of our full model. This demonstrates that our $\mathcal{L}_{SDS}$ with Gaussian Dropout and diffusion timestep annealing further yield clearer details.}
    \label{ablation}
\end{figure}
\begin{figure}
    \centering
    \includegraphics[width=1\textwidth]{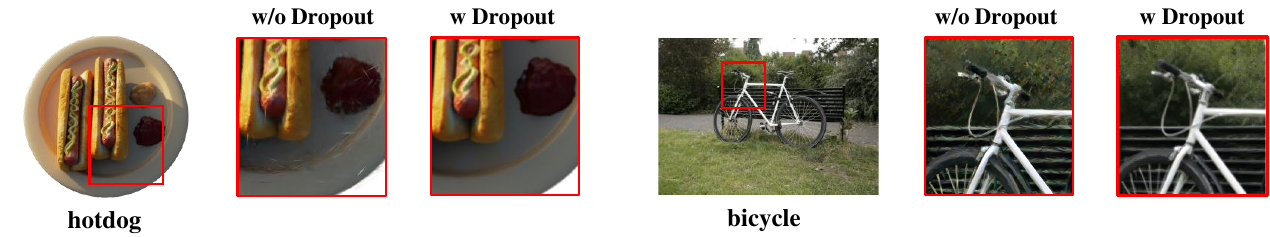}
    \caption{Ablation study on Gaussian Dropout.}
    \label{ablation_dropout}
\end{figure}

\subsection{Ablation Studies}

In Tab.~\ref{table_ablation}, we perform ablation experiments on the components proposed in GaussianSR. Initially, we train the baseline 3DGS model with low-resolution inputs and subsequently render the high-resolution views directly. 
The effectiveness of each proposed component in GaussianSR is evaluated by gradually incorporating them into the model. 
In the second row of Table~\ref{table_ablation}, we optimize the high-resolution 3DGS solely using $\mathcal{L}_{MSE}$. Subsequently, in the third row, we incorporate $\mathcal{L}_{SDS}$ and Gaussian Dropout based on the second row. The effectiveness of the diffusion timestep annealing strategy is evaluated in the last row.
Quantitative results unequivocally demonstrate that our proposed GaussianSR substantially enhances the quality of high-resolution novel views synthesized solely from low-resolution inputs. Additionally, we visualize the renderings of high-resolution novel views to assess the efficacy of our proposed components. As illustrated in Figure~\ref{ablation}, our method effectively mitigates the presence of artifacts that may be present in 3DGS renderings. Furthermore, Figure~\ref{ablation_dropout} showcases results with and without Gaussian Dropout, revealing a notable reduction in redundant Gaussian primitives facilitated by Gaussian Dropout.

\section{Conclusion}
In this paper, we propose GaussianSR, an innovative method for synthesizing high-resolution novel views from low-resolution inputs. 
Our approach is grounded in 3D Gaussian Splatting (3DGS), which offers faster rendering speed.
To address the challenge of limited high-resolution data, we employ Score Distillation Sampling (SDS) to distill generative priors of 2D super-resolution diffusion models. However, the direct application of SDS can lead to redundant Gaussian primitives due to the inherent randomness of generative priors. To mitigate this issue, we propose two straightforward yet effective techniques to reduce stochastic disturbance introduced by SDS. Experimental results demonstrate that GaussianSR excels in synthesizing higher-quality high-resolution novel views.

\paragraph{Limitation and Future Works.}
While our method shows promising results in high-resolution novel view synthesis (HRNVS), there remain limitations to be improved in future work. Our reliance on priors distilled from 2D super-resolution models constrains our performance to the capabilities of the specific 2D models employed. Future improvements could involve distilling priors from multiple 2D super-resolution models trained on diverse datasets, potentially enhancing performance and generalization.

\newpage

\bibliographystyle{plainnat}
% \small

\bibliography{arxiv}

\newpage
\appendix

\section{Discussion of Hyperparameters}
In this section, we discuss the hyperparameters selected in our method, including the weight-balancing parameter $\lambda$ of $\mathcal{L}_{MSE}$ and $\mathcal{L}_{SDS}$ in Sec.~\ref{sup_a1}, the annealing interval $N$ in Sec.~\ref{sup_a2}
and Gaussian Dropout probability $p$ in Sec.~\ref{sup_a3}.

\subsection{Weight-Balancing Parameter \texorpdfstring{$\lambda$}{lambda}}
\label{sup_a1}
To alleviate the shortage of data, we propose to leverage off-the-shelf 2D diffusion priors distilled by $\mathcal{L}_{SDS}$. Meanwhile, to maintain the consistency of low-resolution views and to prevent color shifts occasionally caused by diffusion model, we take $\mathcal{L}_{MSE}$ into consideration as a regularizer. Then, the high-resolution 3DGS is optimized by $\mathcal{L}_{MSE} + \lambda \mathcal{L}_{SDS}$, achieving high-resolution novel view synthesis. In this section, we make an analysis for the wight-balancing parameter $\lambda$. We randomly select three scenes form Blender dataset, Mip-NeRF 360 dataset and Deep Blending dataset to evaluate the performance under different $\lambda$. Tab.~\ref{tab_sup_lambda} presents that the best performance across three views in terms of PSNR and SSIM  can be attained when $\lambda = 0.001$, whereas GaussianSR performs best in LPIPS when $\lambda$ is set to other values in "stump" and "playroom". After evaluating all aspects, we chose $\lambda = 0.001$ for training.

\subsection{Annealing Interval \texorpdfstring{$N$}{N}}
\label{sup_a2}
In order to reduce the randomness brought by generative priors, we shrink the diffusion timestep sampling range by an annealing strategy. In this section, we conduct an ablation of the annealing interval $N$ in Eq.~\ref{eq_t}. We evaluate the qualitative results under different $N$ on three scenes randomly selected from Mip-NeRF 360 \cite{barron2022mip} and Deep Blending \cite{hedman2018deep} dataset. As shown in Tab.~\ref{tab_sup_n}, GaussianSR achieves higher PSNR in the three scenes when $N$ is set to 100. Therefore, we shrink the diffusion timestep range every 100 iterations during training.

\subsection{Gaussian Dropout Probability \texorpdfstring{$p$}{p}}
\label{sup_a3}
\begin{figure}[h]
    \centering
    \includegraphics[width=1\textwidth]{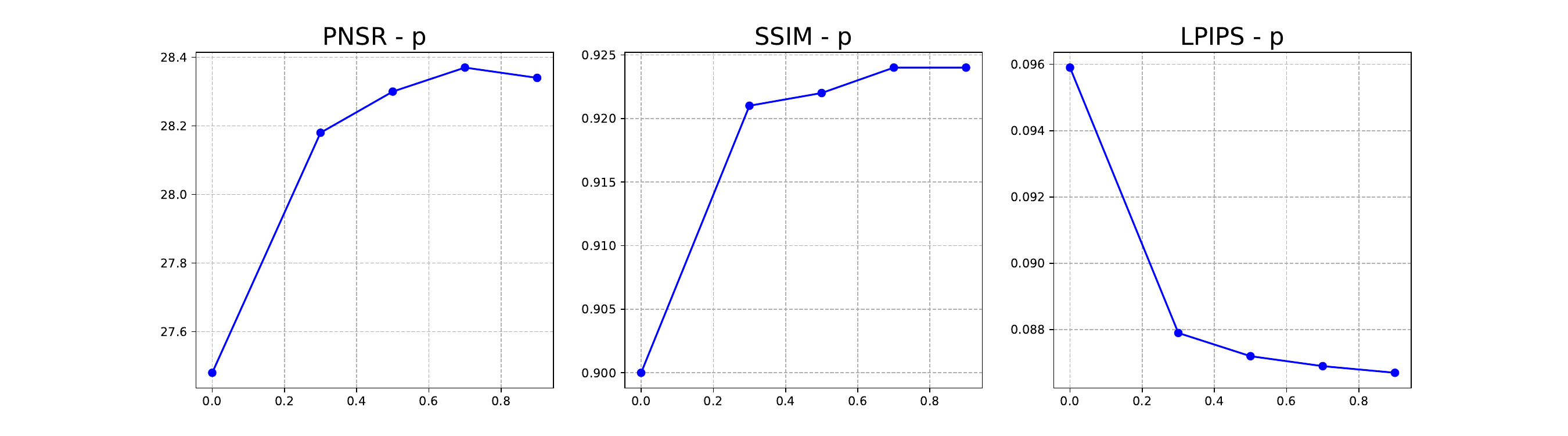}
    \caption{Ablation studies of Gaussian Dropout probability $p$ on Blender dataset.}
    \label{sup_p}
\end{figure}

As described in Eq.~\ref{eq_dropout}, we utilize the certain probability $p$ to generate a mask for suppress the cloning and splitting of some Gaussian primitives. Therefore, the performance is heavily related to the probability $p$. To chosen the $p$ with higher performance, we conduct an ablation under different dropout probabilities $p$ on the Blender dataset \cite{mildenhall2021nerf}.  
Referring to Fig.~\ref{sup_p}, GaussianSR demonstrates the best performance in terms of PSNR when $p = 0.7$, whereas it performs best in LPIPS when $p = 0.9$. Taking all aspects into consideration, $p = 0.7$ is chosen in the training process.

\section{Additional Results}
\begin{figure}[h]
    \centering
    \includegraphics[width=1\textwidth]{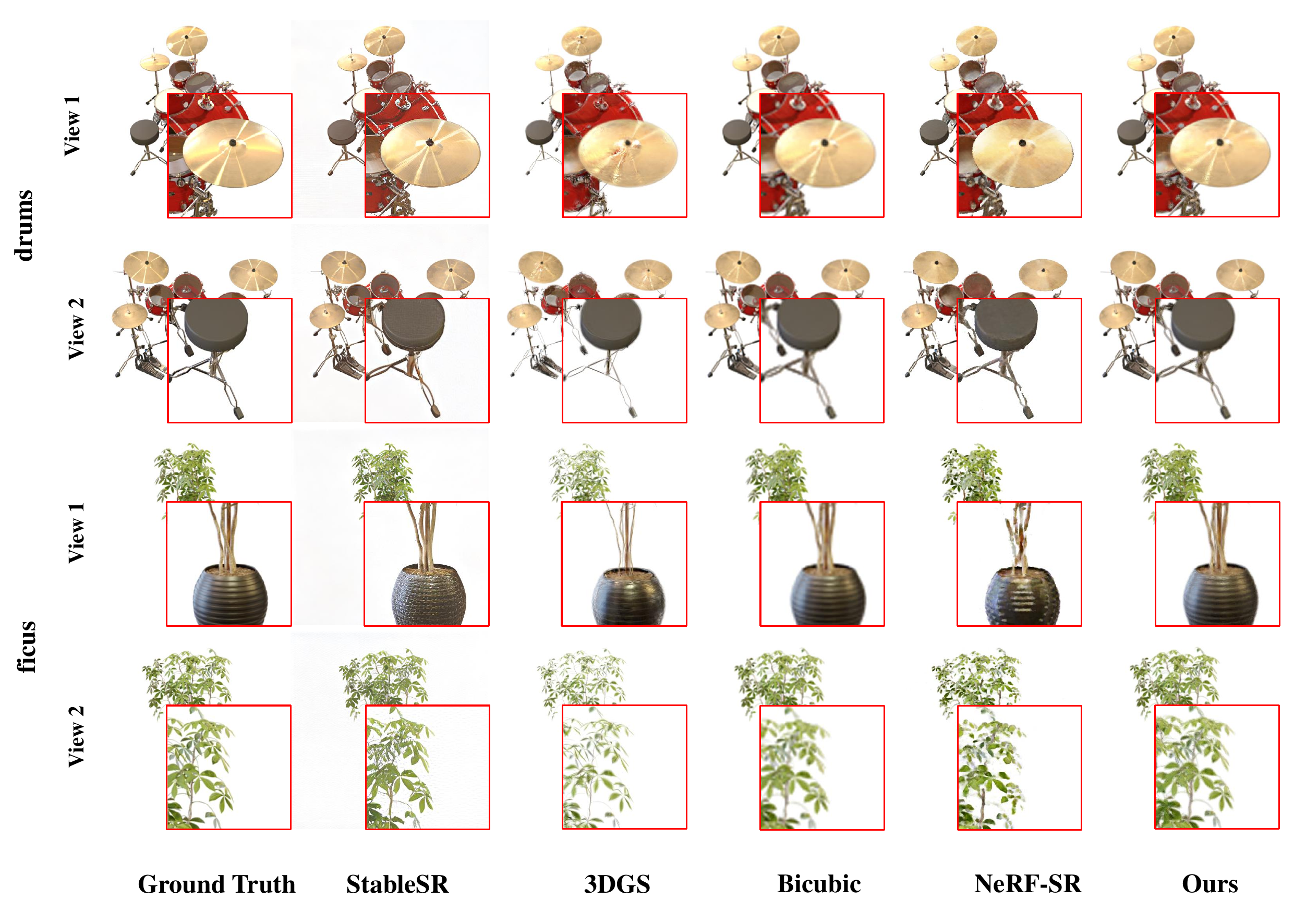}
    \caption{Qualitative comparison of our method with vanalia 3DGS, bicubic interpolation, NeRF-SR and
StableSR on Blender dataset for the HRNVS (×4).}
    \label{sup_result_blender}
\end{figure}

\begin{figure}
    \centering
    \includegraphics[width=1\textwidth]{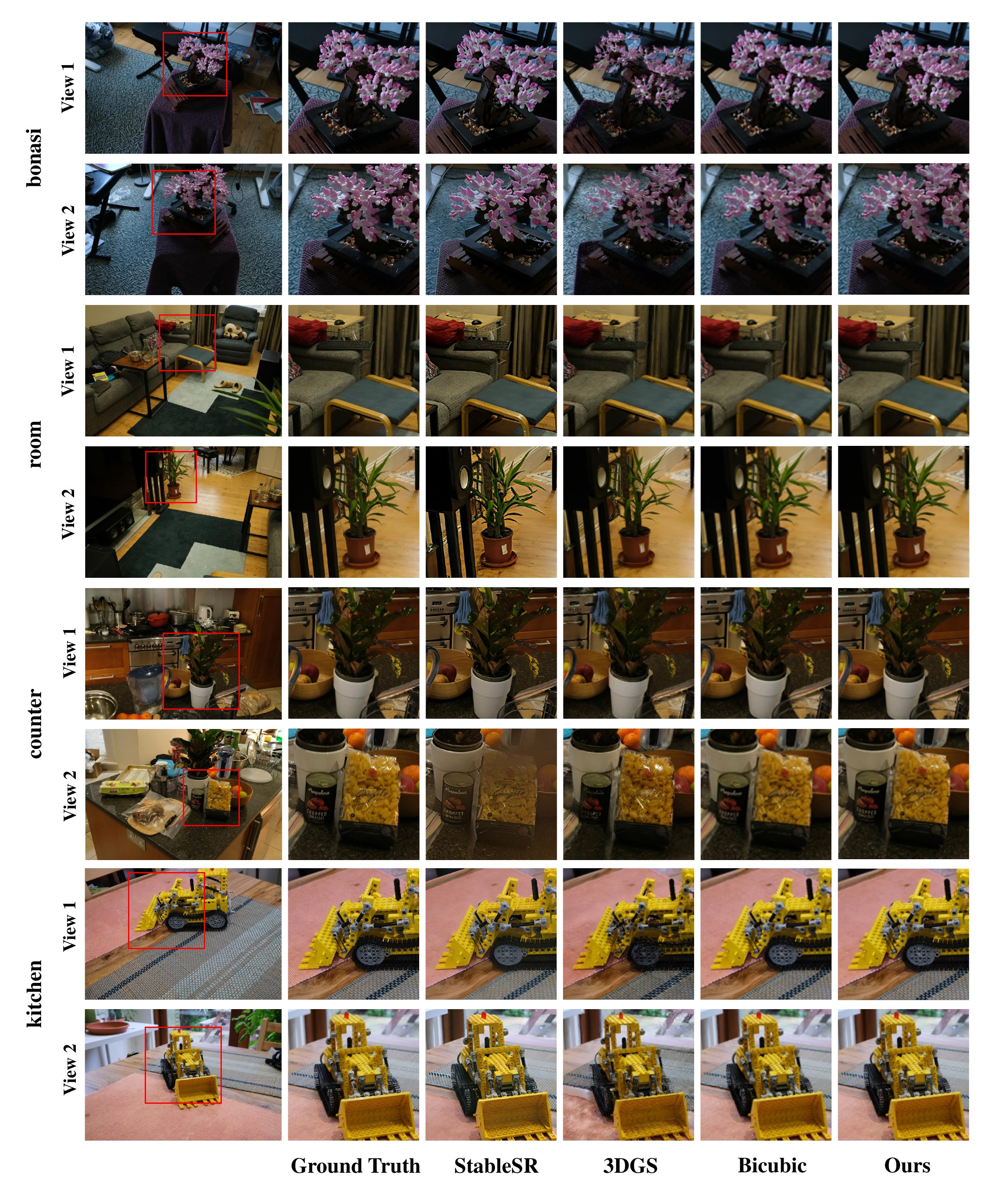}
    \caption{Qualitative comparison of our method with vanalia 3DGS, bicubic interpolation and
StableSR in the indoor scenes of Mip-NeRF 360 dataset for the HRNVS (×4). The results are zoom-in version of the red box region.}
    \label{sup_results_360_indoor}
\end{figure}

In this section, we provide more qualitative and quantitative results on Blender dataset \cite{mildenhall2021nerf}, Mip-NeRF 360 dataset \cite{barron2022mip}, and Deep Blending dataset \cite{hedman2018deep}.
% We evaluate GaussianSR with $\times 4$ High-Resolution Novel View Synthesis (HRNVS) in Blender dataset \cite{mildenhall2021nerf}.
For Blender dataset, Tab.~\ref{per_scene_blender} presents per-scene metrics for $\times 4$ HRNVS. For each scene, we calculate the arithmetic mean of each metric averaged over all test views. 
More qualitative comparison on Blender dataset against leading methods is shown in Fig.~\ref{sup_result_blender}.
And per-scene metrics for $\times 4$ HRNVS on Mip-NeRF 360 dataset are shown in Tab.~\ref{per_scene_360}, which demonstrates that our GaussianSR has the ability to synthesize higher-quality high-resolution novel views in most scenes. Following 3DGS \cite{kerbl3Dgaussians}, we select a subset of Deep Blending dataset to evaluate our method, where "drjohnson" and "playroom" are chosen. And the per-scene metrics of "drjohnson" and "playroom" are compiled in Tab.~\ref{per_scene_db}. Furthermore, more qualitative evaluation on Mip-NeRF 360 and Deep Blending dataset are presented in Fig.~\ref{sup_results_360_indoor}, Fig.~\ref{sup_results_360_outdoor} and Fig.~\ref{sup_results_db}. 
We also provide the video of our results in the supplementary materials which can entirely show the strength of our method.

% \newpage
\begin{table}[h]
\centering
\caption{Ablation studies for weight-balancing parameter $\lambda$ on Blender dataset, Mip-NeRF 360 dataset and Deep Blending dataset.}
\begin{tabularx}{\textwidth}{l|*{3}{>{\centering\arraybackslash}X}|*{3}{>{\centering\arraybackslash}X}|*{3}{>{\centering\arraybackslash}X}}

     & \multicolumn{3}{c|}{\textit{ficus}} & \multicolumn{3}{c|}{\textit{stump}}& \multicolumn{3}{c}{\textit{playroom}} \\

 $\lambda$ & PSNR$\uparrow$ & SSIM$\uparrow$ & LPIPS$\downarrow$ & PSNR$\uparrow$ & SSIM$\uparrow$ & LPIPS$\downarrow$& PSNR$\uparrow$ & SSIM$\uparrow$ & LPIPS$\downarrow$ \\
\hline
1 &24.47  &0.927 & 0.071 &24.31 &\cellcolor{pink!50}0.571 &0.402 & 25.10 & 0.834 & 0.362\\

 0.1&24.37  &0.927 & 0.071 &19.70 &0.372 &0.621 & \cellcolor{pink!50}28.58 & \cellcolor{pink!50}0.878 & \cellcolor{red!50}0.308\\

0.01 &24.35  &0.927 & 0.071 &\cellcolor{pink!50}24.33  &\cellcolor{pink!50}0.571 &\cellcolor{pink!50}0.402& 26.83 & 0.853 & 0.353\\

0.001 & \cellcolor{red!50}29.19 &\cellcolor{red!50}0.952 & \cellcolor{red!50}0.052 
&\cellcolor{red!50}24.38  &\cellcolor{red!50}0.574  &0.408 & \cellcolor{red!50}28.76 & \cellcolor{red!50}0.881 & \cellcolor{pink!50}0.309\\

0.0001 & \cellcolor{pink!50}28.73  & \cellcolor{pink!50}0.948 & \cellcolor{pink!50}0.056 &24.28  &\cellcolor{pink!50}0.571  &\cellcolor{red!50}0.401 & 25.20 & 0.835 &0.362\\

% \hline
\end{tabularx}
\label{tab_sup_lambda}
\end{table}

\begin{table}[h]
\centering
\caption{Ablation studies for annealing interval $N$ on Mip-NeRF 360 dataset and Deep Blending dataset .}
\begin{tabularx}{\textwidth}{l|*{3}{>{\centering\arraybackslash}X}|*{3}{>{\centering\arraybackslash}X}|*{3}{>{\centering\arraybackslash}X}}

 & \multicolumn{3}{c|}{\textit{kitchen}} & \multicolumn{3}{c|}{\textit{treehill}}& \multicolumn{3}{c}{\textit{drjohnson}} \\

 $N$ & PSNR$\uparrow$ & SSIM$\uparrow$ & LPIPS$\downarrow$ & PSNR$\uparrow$ & SSIM$\uparrow$ & LPIPS$\downarrow$& PSNR$\uparrow$ & SSIM$\uparrow$ & LPIPS$\downarrow$ \\
\hline
500 & \cellcolor{pink!50}28.19 & 0.790 & 0.282 & 21.80 & 0.482 & \cellcolor{red!50}0.473 & 27.79 & 0.864 &\cellcolor{pink!50}0.307 \\

300 & 28.02 & \cellcolor{pink!50}0.791 & \cellcolor{pink!50}0.278 & \cellcolor{pink!50}21.86 &\cellcolor{pink!50} 0.485 & \cellcolor{red!50}0.473 &27.80 &\cellcolor{red!50}0.866 & \cellcolor{red!50}0.305  \\

100 & \cellcolor{red!50}28.27  &\cellcolor{red!50}0.794 & \cellcolor{red!50}0.276 &\cellcolor{red!50}21.98  &\cellcolor{red!50}0.490  &\cellcolor{pink!50}0.478 & \cellcolor{red!50}27.81 & \cellcolor{pink!50}0.865 & \cellcolor{pink!50}0.307\\

% \hline
\end{tabularx}
\label{tab_sup_n}
\end{table}

\clearpage
\newpage
\begin{table}[h]
\centering
\caption{Quantitative evaluation for HRNVS ($\times4$) on the Blender dataset. For each scene, we report the arithmetic mean of each metric averaged over all test views.}

\vspace{0.3cm}

\begin{subtable}[htbp]{\textwidth}
    \centering
    \newcolumntype{Y}{>{\raggedleft\arraybackslash}X} 
	\newcolumntype{Z}{>{\centering\arraybackslash}X} 
        \centerline{\textbf{PSNR}}
        \begin{tabularx}{\textwidth}        {p{0.14\textwidth}|ZZZZZZZZ|Z}
            % \hline
            Method & \textit{chair} & \textit{drums} &\textit{ficus} & \textit{hotdog} & \textit{lego} &  \textit{materials} &\textit{mic} &\textit{ship} & \textit{Average}\\
            \hline
            3DGS \cite{kerbl3Dgaussians} & 24.14 & 19.33 & 21.72 & 25.97 & 20.46 & 20.27 & 21.44 & 20.90 & 21.78 \\
            StableSR \cite{wang2023exploiting} & 24.69 & 21.12 & 23.45 & 26.68 & 22.64 & 22.98 & 24.88&22.17 & 23.58 \\
            Bicubic & 28.81 & 23.35 & 27.32 & 31.67 & 27.19 & 26.07 & 27.90 & 25.51 & 27.23 \\
                NeRF-SR \cite{wang2022nerf} & \cellcolor{red!50}{30.18} & \cellcolor{pink!50}23.50 & \cellcolor{pink!50}22.72 & \cellcolor{red!50}{34.38} & \cellcolor{red!50}{29.21} & \cellcolor{red!50}{28.08} & \cellcolor{pink!50}27.25 & \cellcolor{red!50}{26.59} & \cellcolor{pink!50}27.81 \\
                Ours  & \cellcolor{pink!50}29.81 & \cellcolor{red!50}{24.05} & \cellcolor{red!50}{29.19} & \cellcolor{pink!50}32.80& \cellcolor{pink!50}28.66 & \cellcolor{pink!50}27.02 & \cellcolor{red!50}{29.15} & \cellcolor{pink!50}26.28 & \cellcolor{red!50}{28.37} \\
            
            % \hline
        \end{tabularx}
\end{subtable}

\vspace{0.5cm}

\begin{subtable}[htbp]{\textwidth}
    \centering
    \newcolumntype{Y}{>{\raggedleft\arraybackslash}X} 
	\newcolumntype{Z}{>{\centering\arraybackslash}X} 
        \centerline{\textbf{SSIM}}
        \begin{tabularx}{\textwidth}        {p{0.14\textwidth}|ZZZZZZZZ|Z}
            % \hline
            Method & \textit{chair} & \textit{drums} &\textit{ficus} & \textit{hotdog} & \textit{lego} &  \textit{materials} &\textit{mic} &\textit{ship} & \textit{Average}\\
            \hline
            3DGS \cite{kerbl3Dgaussians} & 0.886 & 0.852 & 0.916 & 0.920 & 0.821 & 0.859 & 0.914 & 0.778 & 0.868 \\
            StableSR \cite{wang2023exploiting} & 0.870 & 0.859 & 0.893 & 0.890 & 0.808 & 0.856 & 0.916 & 0.740 & 0.854 \\
            Bicubic & 0.918 & 0.892 &\cellcolor{pink!50} 0.934 & 0.951 & 0.902 & 0.916 & \cellcolor{pink!50}0.947 & 0.824 & 0.911 \\
                NeRF-SR \cite{wang2022nerf} & \cellcolor{red!50}{0.937} & \cellcolor{pink!50}0.903 & 0.904 & \cellcolor{red!50}{0.964} & \cellcolor{red!50}{0.931} & \cellcolor{red!50}{0.932} & 0.943 & \cellcolor{pink!50}{0.836} & \cellcolor{pink!50}0.919 \\
                Ours  & \cellcolor{pink!50}0.931 & \cellcolor{red!50}{0.909} & \cellcolor{red!50}{0.952} & \cellcolor{pink!50}0.959& \cellcolor{pink!50}0.924 & \cellcolor{pink!50}0.925 & \cellcolor{red!50}{0.958} & \cellcolor{red!50}0.837 & \cellcolor{red!50}{0.924} \\
            
            % \hline
        \end{tabularx}
\end{subtable}

\vspace{0.5cm}

\begin{subtable}[htbp]{\textwidth}
    \centering
    \newcolumntype{Y}{>{\raggedleft\arraybackslash}X} 
	\newcolumntype{Z}{>{\centering\arraybackslash}X} 
        \centerline{\textbf{LPIPS}}
        \begin{tabularx}{\textwidth}        {p{0.14\textwidth}|ZZZZZZZZ|Z}
            % \hline
            Method & \textit{chair} & \textit{drums} &\textit{ficus} & \textit{hotdog} & \textit{lego} &  \textit{materials} &\textit{mic} &\textit{ship} & \textit{Average}\\
            \hline
            3DGS \cite{kerbl3Dgaussians} & 0.083 & \cellcolor{pink!50}0.108 & \cellcolor{pink!50}0.062 & 0.077 & 0.142 & 0.107 & 0.061 & \cellcolor{pink!50}0.192 & 0.104 \\
            StableSR \cite{wang2023exploiting} & 0.178 & 0.178 & 0.201 & 0.200 & 0.209 & 0.227 & 0.197&0.265 & 0.207 \\
            Bicubic & 0.091& 0.118 & 0.073 & 0.078 & 0.132 & 0.096 & \cellcolor{pink!50}0.063 & 0.219 & 0.115 \\
                NeRF-SR \cite{wang2022nerf} & \cellcolor{red!50}0.068 & \cellcolor{pink!50}0.108 & 0.100 & \cellcolor{red!50}{0.053} & \cellcolor{red!50}{0.090} & \cellcolor{red!50}{0.076} & 0.078 & {0.198} & \cellcolor{pink!50}0.097 \\
                Ours  & \cellcolor{pink!50}0.074 & \cellcolor{red!50}{0.095} & \cellcolor{red!50}{0.052} & \cellcolor{pink!50}0.061& \cellcolor{pink!50}0.106 & \cellcolor{pink!50}0.083 & \cellcolor{red!50}{0.050} & \cellcolor{red!50}0.175 & \cellcolor{red!50}{0.087} \\
            
            % \hline
        \end{tabularx}
\end{subtable}

\label{per_scene_blender}
\end{table}

\begin{table}[h]
\centering
\caption{Quantitative evaluation for HRNVS ($\times4$) on the Mip-NeRF 360 dataset. For each scene, we report the arithmetic mean of each metric averaged over all test views.}
\vspace{0.3cm}
\begin{subtable}[htbp]{\textwidth}
    \centering
    \newcolumntype{Y}{>{\raggedleft\arraybackslash}X} 
	\newcolumntype{Z}{>{\centering\arraybackslash}X} 
        \centerline{\textbf{PSNR}}
        \begin{tabularx}{\textwidth}        {p{0.14\textwidth}|ZZZZZ|ZZZZ}
            % \hline
            Method & \textit{bicycle} & \textit{flowers} &\textit{garden} & \textit{stump} & \textit{treehill} &  \textit{bonasi} &\textit{counter} &\textit{kitchen} & \textit{room}\\
            \hline
            3DGS \cite{kerbl3Dgaussians} & 18.48 & 15.80 & 19.73 & 18.65&15.91 & 23.18 & 23.52 & 20.66 & 26.58 \\
            StableSR \cite{wang2023exploiting} & 20.61 & 18.28 & 20.77 & 21.83 & 20.23 & 23.32 & 23.33 &22.43 & 24.61 \\
            
            Bicubic & \cellcolor{pink!50}22.30 & \cellcolor{pink!50}20.43 & \cellcolor{pink!50}23.27 & \cellcolor{pink!50}24.20 & \cellcolor{red!50}22.05 & \cellcolor{pink!50}29.43 & \cellcolor{pink!50}27.38 & \cellcolor{pink!50}27.40 & \cellcolor{pink!50}29.78 \\
                Ours  & \cellcolor{red!50}22.63 & \cellcolor{red!50}{20.61} & \cellcolor{red!50}{23.72} & \cellcolor{red!50}24.38& \cellcolor{pink!50}21.98 & \cellcolor{red!50}30.23 & \cellcolor{red!50}{28.02} & \cellcolor{red!50}28.27 & \cellcolor{red!50}{30.58} \\
            
            % \hline
        \end{tabularx}
\end{subtable}

\vspace{0.5cm}

\begin{subtable}[htbp]{\textwidth}
    \centering
    \newcolumntype{Y}{>{\raggedleft\arraybackslash}X} 
	\newcolumntype{Z}{>{\centering\arraybackslash}X} 
        \centerline{\textbf{SSIM}}
        \begin{tabularx}{\textwidth}        {p{0.14\textwidth}|ZZZZZ|ZZZZ}
            % \hline
            Method & \textit{bicycle} & \textit{flowers} &\textit{garden} & \textit{stump} & \textit{treehill} &  \textit{bonasi} &\textit{counter} &\textit{kitchen} & \textit{room}\\
            \hline
            3DGS \cite{kerbl3Dgaussians} & 0.405 & 0.326 & 0.463 & 0.405 & 0.404 & 0.773 & 0.754 & 0.694 & 0.826 \\
            StableSR \cite{wang2023exploiting} & 0.337 & 0.315 & 0.398 & 0.409 &  0.384& 0.651 & 0.656 & 0.546 & 0.678 \\
            Bicubic & \cellcolor{pink!50}0.491 & \cellcolor{pink!50}0.436 & \cellcolor{pink!50}0.516 & \cellcolor{pink!50}0.567 &\cellcolor{pink!50}0.489& \cellcolor{pink!50}0.871 & \cellcolor{pink!50}0.823 & \cellcolor{pink!50}0.759 & \cellcolor{pink!50}0.861 \\
                Ours  & \cellcolor{red!50}0.511 & \cellcolor{red!50}{0.453} & \cellcolor{red!50}{0.556} & \cellcolor{red!50}0.574& \cellcolor{red!50}0.490 & \cellcolor{red!50}0.883 & \cellcolor{red!50}{0.839} & \cellcolor{red!50}0.794 & \cellcolor{red!50}{0.873} \\
            
            % \hline
        \end{tabularx}
\end{subtable}

\vspace{0.5cm}

\begin{subtable}[htbp]{\textwidth}
    \centering
    \newcolumntype{Y}{>{\raggedleft\arraybackslash}X} 
	\newcolumntype{Z}{>{\centering\arraybackslash}X} 
        \centerline{\textbf{LPIPS}}
        \begin{tabularx}{\textwidth}        {p{0.14\textwidth}|ZZZZZ|ZZZZ}
            % \hline
            Method & \textit{bicycle} & \textit{flowers} &\textit{garden} & \textit{stump} & \textit{treehill} &  \textit{bonasi} &\textit{counter} &\textit{kitchen} & \textit{room}\\
            \hline
            3DGS \cite{kerbl3Dgaussians} & 0.481 & 0.509 & 0.446 & 0.476 & 0.526 & 0.353 & 0.330 & 0.336 & \cellcolor{pink!50}0.322 \\
            StableSR \cite{wang2023exploiting} & \cellcolor{red!50}0.385 & \cellcolor{red!50}0.440 & \cellcolor{red!50}0.374 & \cellcolor{pink!50}0.424 & \cellcolor{red!50}0.415 & 0.360 & 0.340 &0.373 & 0.336 \\
            Bicubic & 0.486 & 0.509 & 0.455 & 0.442 & 0.512 & \cellcolor{pink!50}0.292 & \cellcolor{pink!50}0.318 & \cellcolor{pink!50}0.313 & 0.327 \\
                Ours  & \cellcolor{pink!50}0.450 & \cellcolor{pink!50}{0.448} & \cellcolor{pink!50}{0.395} & \cellcolor{red!50}0.408& \cellcolor{pink!50}0.478 & \cellcolor{red!50}0.273 & \cellcolor{red!50}{0.283} & \cellcolor{red!50}0.276 & \cellcolor{red!50}{0.303} \\
            
            % \hline
        \end{tabularx}
\end{subtable}

\label{per_scene_360}
\end{table}

\begin{table}[htbp]
\centering
\caption{Quantitative evaluation for HRNVS ($\times4$) on the Deep Blending dataset. For each scene, we report the arithmetic mean of each metric averaged over all test views.}
% \vspace{0.3cm}
\begin{subtable}[htbp]{\textwidth}
    \centering 
        \centerline{\textbf{PSNR}}
        \vspace{1mm}
        \begin{tabular}{l|cc|c}
            % \hline
            Method & \textit{drjohnson} & \textit{playroom} &\textit{Average} \\
            \hline
            3DGS \cite{kerbl3Dgaussians} & 26.64 & 27.17 & 26.64  \\
            StableSR \cite{wang2023exploiting} & 24.02 & 23.83 & 23.93  \\
            
            Bicubic & \cellcolor{pink!50}27.73 & \cellcolor{pink!50}28.29 & \cellcolor{pink!50}28.00 \\
                Ours  & \cellcolor{red!50}27.81 & \cellcolor{red!50}{28.76} & \cellcolor{red!50}{28.28}  \\
        \end{tabular}
\end{subtable}

% \vspace{0.1cm}

\begin{subtable}[htbp]{\textwidth}
    \centering 
        \centerline{\textbf{SSIM}}
        \vspace{1mm}
        \begin{tabular}{l|cc|c}
            % \hline
            Method &\textit{drjohnson} & \textit{playroom} &\textit{Average} \\
            \hline
            3DGS \cite{kerbl3Dgaussians} & 0.847 & 0.860 & 0.853  \\
            StableSR \cite{wang2023exploiting} & 0.706 & 0.710 & 0.708  \\
            
            Bicubic & \cellcolor{pink!50}0.858 &\cellcolor{pink!50}0.870 & \cellcolor{pink!50}0.864 \\
                Ours  & \cellcolor{red!50}0.865 & \cellcolor{red!50}{0.881} & \cellcolor{red!50}{0.873}  \\
        \end{tabular}
\end{subtable}

% \vspace{0.1cm}

\begin{subtable}[htbp]{\textwidth}
    \centering 
        \centerline{\textbf{LPIPS}}
        \vspace{1mm}
        \begin{tabular}{l|cc|c}
            % \hline
            Method & \textit{drjohnson} & \textit{playroom} &\textit{Average} \\
            \hline
            3DGS \cite{kerbl3Dgaussians} & \cellcolor{pink!50}0.313 & \cellcolor{pink!50}0.312 & \cellcolor{pink!50}0.312  \\
            StableSR \cite{wang2023exploiting} & 0.332 & 0.318 & 0.325  \\
            
            Bicubic & 0.329 & 0.330 & 0.330 \\
                Ours  & \cellcolor{red!50}0.307 & \cellcolor{red!50}{0.309} & \cellcolor{red!50}{0.308}  \\
        \end{tabular}
\end{subtable}

\label{per_scene_db}
\end{table}

\begin{figure}
    \centering
    \includegraphics[width=1\textwidth]{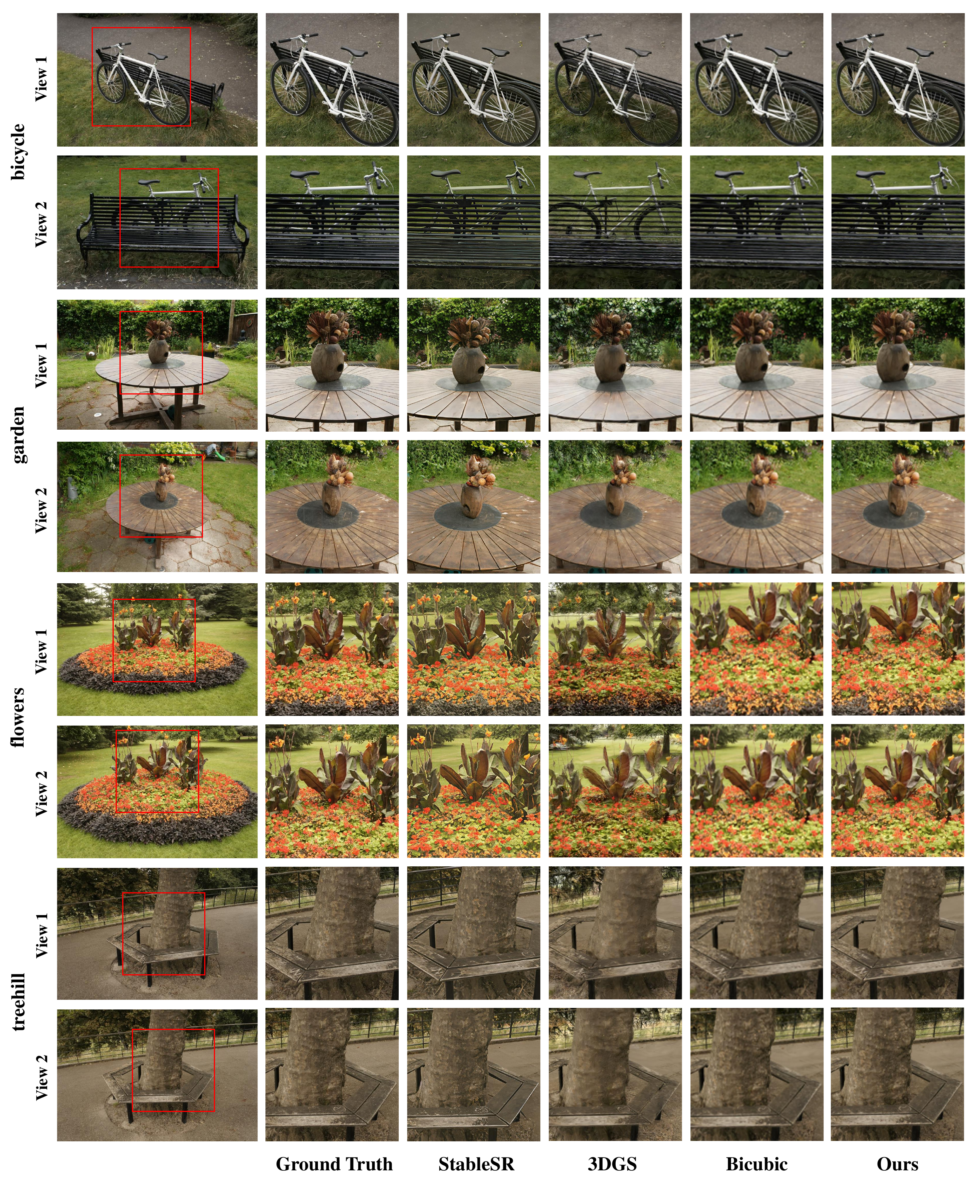}
    \caption{Qualitative comparison of our method with vanalia 3DGS, bicubic interpolation and
StableSR in the outdoor scenes of Mip-NeRF 360 dataset for the HRNVS (×4). The results are zoom-in version of the red box region.}
    \label{sup_results_360_outdoor}
\end{figure}

\begin{figure}
    \centering
    \includegraphics[width=1\textwidth]{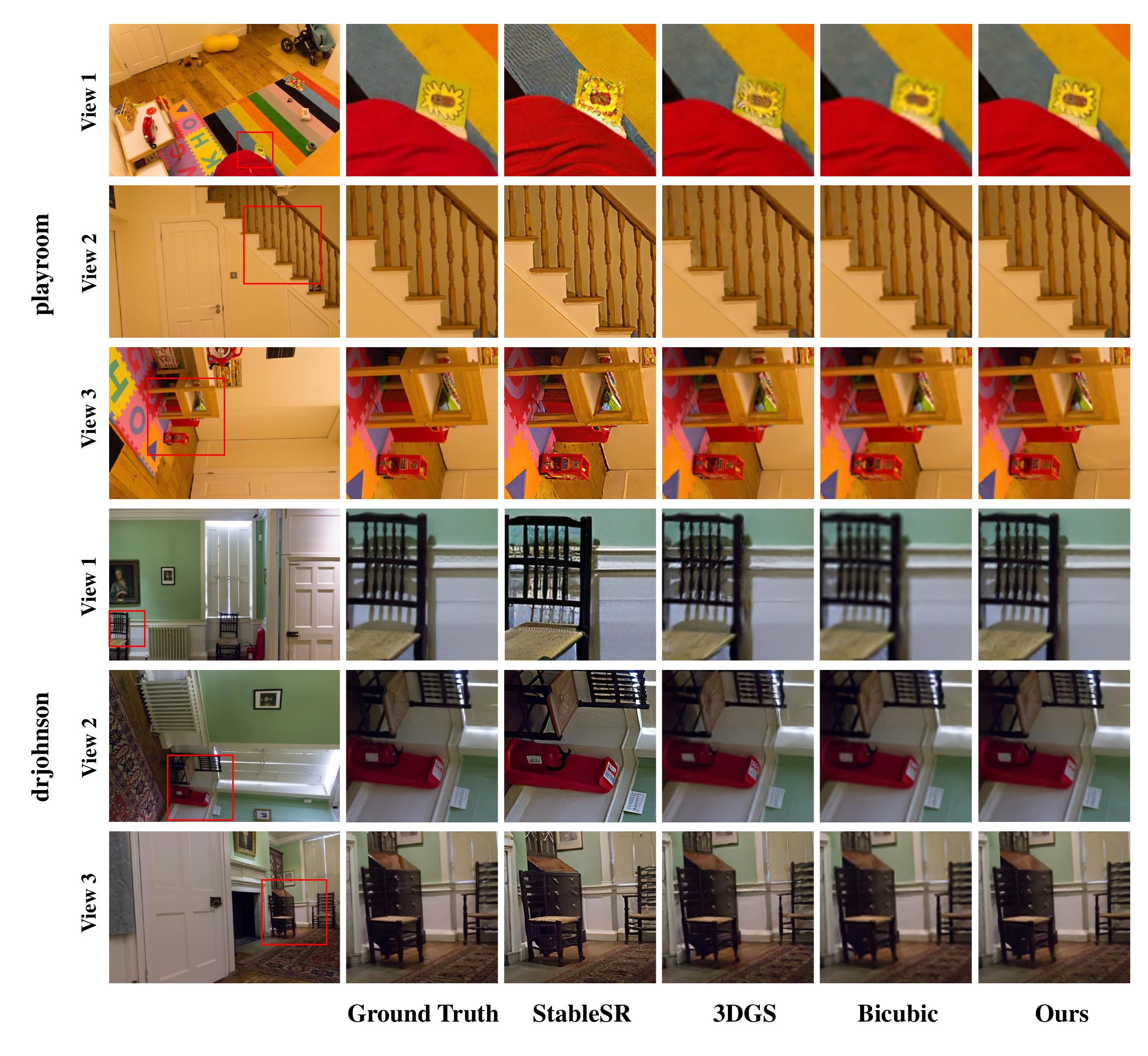}
    \caption{Qualitative comparison of our method with vanalia 3DGS, bicubic interpolation and
StableSR on Deep Blending dataset for the HRNVS (×4). The results are zoom-in version of the red box region.}
    \label{sup_results_db}
\end{figure}

\end{document}